\documentclass[runningheads]{llncs}
\usepackage{graphicx}
\usepackage{subfig}
\usepackage{multirow}
\usepackage{pdfpages}
\usepackage[colorlinks=true]{hyperref}
\usepackage[ruled,vlined, linesnumbered]{algorithm2e}

\makeatletter
\newcommand\footnoteref[1]{\protected@xdef\@thefnmark{\ref{#1}}\@footnotemark}
\makeatother

\newcommand{\comment}[1]{}
\newcommand{\mM}{{\mathcal M}}

\usepackage{hyperref}

\usepackage{amsmath}
\usepackage{amssymb}
\usepackage{caption}
\usepackage{wrapfig}
\usepackage{bm}
\usepackage{footnote}
\makesavenoteenv{tabular}
\makesavenoteenv{table}
\usepackage{pifont}
\newcommand{\cmark}{\ding{51}}%
\newcommand{\xmark}{\ding{55}}%
\newcommand{\printfnsymbol}[1]{%
  \textsuperscript{\@*}%
  }
  
\newcommand{\ubold}[1]{\fontseries{b}\selectfont#1}
\usepackage[export]{adjustbox}
\usepackage{array}
\usepackage{arydshln}
\usepackage{booktabs}
\newcommand{\ra}[1]{\renewcommand{\arraystretch}{#1}}
\begin{document}
\title{STRUDEL: Self-Training with Uncertainty Dependent Label Refinement across Domains}


%
\titlerunning{STRUDEL: Self-Training with Uncertainty Dependent Label Refinement}
%

\author{Fabian Gröger \inst{1}\inst{2} \and Anne-Marie Rickmann \inst{1} \and
Christian Wachinger \inst{1} }

\authorrunning{Gröger et al.}
%
\institute{
Artificial Intelligence in Medical Imaging (AI-Med), KJP, LMU M\"unchen, Germany \\
 \and
Computer Aided Medical Procedures, Technische Universität München, Germany}
\maketitle              
\begin{abstract}


We propose an unsupervised domain adaptation (UDA) approach for white matter hyperintensity (WMH) segmentation, which uses Self-TRaining with Uncertainty DEpendent Label refinement (STRUDEL). 
Self-training has recently been introduced as a highly effective method for UDA, which is based on self-generated pseudo labels. 
However, pseudo labels can be very noisy and therefore deteriorate model performance. 
We propose to predict the uncertainty of pseudo labels and integrate it in the training process with an uncertainty-guided loss function to highlight labels with high certainty. 
STRUDEL is further improved by incorporating the segmentation output of an existing method in the pseudo label generation that showed high robustness for WMH segmentation. 
In our experiments, we evaluate STRUDEL with a standard U-Net and a modified network with a higher receptive field. 
Our results on WMH segmentation across datasets demonstrate the significant improvement of STRUDEL with respect to standard self-training. 


\end{abstract}
\section{Introduction}
\begin{figure}[t]
 \centering
 \includegraphics[keepaspectratio,width=\textwidth,height=\textheight]{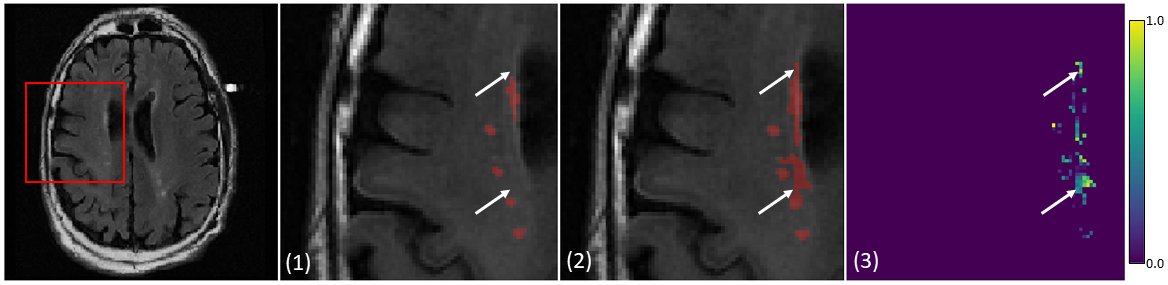}
 \caption{Illustration of a FLAIR scan with (1) ground truth WMH, (2) noisy pseudo labels, and  (3)  corresponding uncertainty map. White arrows point to false positive predictions with higher uncertainty values (brighter pixels).}
 \label{fig:lesionViz}
\end{figure}

Dementia presents a highly relevant societal challenge due to the ever-aging population. 
Research shows that aging-related structural and functional changes in the brain may manifest as cerebral small vessel disease (SVD), which is a major contributor to the risk of developing dementia~\cite{limor2015}.
A promising neuroimaging biomarker for SVD are white matter hyperintensities (WMHs) of presumed vascular origin. 
WMHs are visible in fluid-attenuated inversion recovery (FLAIR) magnetic resonance imaging (MRI)  as diffuse regions of brighter intensity than surrounding white matter~\cite{prins2015}.
\noindent
Convolutional neural networks (CNNs) have achieved remarkable performances for WMH segmentation~\cite{kuijf2019standardized}. 
However, CNNs are highly dependent on the training set and the performance can steeply decrease on a  target domain with a large domain shift. 
In a recent comparison, this resulted in traditional segmentation software producing higher quality WMH labels than CNNs~\cite{vanderbecq2020}. 
Unsupervised domain adaptation (UDA) attempts to overcome the problem of domain shift without using target data annotations.
Not relying on target annotations is a key benefit, as medical annotations are typically scarce and therefore will not cover the wide range of acquisition protocols, scanner types, artifacts, or patient statistics that make up domain differences in MRI. 
Self-training is a recent approach for UDA, where a segmentation model is first trained on annotated source data and then applied on target data to infer pseudo labels.
These self-generated pseudo labels are then integrated into the network training to achieve the domain adaptation. 
However, pseudo labels tend to be noisy, as illustrated in Fig.~\ref{fig:lesionViz}, which necessitates estimating the reliability of pseudo labels to avoid propagating label errors. 

\noindent
In this work, we propose STRUDEL, a Self-TRaining approach with Uncertainty DEpendent Label refinement. 
It is motivated by earlier work on brain lesion segmentation~\cite{nair2018}, which demonstrated that uncertainty measures are an indicator for erroneous pixel-wise predictions. 
Following a Bayesian segmentation approach, we estimate the uncertainty for pseudo labels, see Fig.~\ref{fig:lesionViz}, and then integrate it in successive model refinements with an uncertainty-guided loss function. 
To further improve the initial pseudo label generation in STRUDEL, we propose to integrate the output of the lesion prediction algorithm (LPA)~\cite{schmidt2017bayesian}, which was reported to achieve robust results across domains~\cite{vanderbecq2020}. 
In our experiments, we evaluate STRUDEL with a  U-Net as the backbone and a modified network with a higher receptive field. 
Our results on WMH segmentation across datasets demonstrate the necessity for domain adaptation and further the significant improvement for integrating uncertainty and LPA in the training process. 


\subsection{Related work}
\textit{White Matter Hyperintensity Segmentation} methods have recently been assessed in the WMH challenge~\cite{kuijf2019standardized}. All top-ranking methods were deep-learning-based and some have achieved superior performance to human observers. Specific inter-scanner robustness experiments showed there is still a need for improving the robustness of these methods which coincides with the findings in~\cite{vanderbecq2020}.

\noindent
\textit{Unsupervised Domain Adaptation} (UDA) approaches transfer a model from a source domain without direct supervision on the target domain and are commonly based on adversarial learning.
These types of methods attempt to learn domain invariant features by minimizing the discrepancy between source and target domain or convert images from one domain to the other.
Different approaches have demonstrated their effectiveness for medical applications~\cite{kamnitsas2017,huo2018adversarial}.
However, the training process for adversarial networks can be a multi-faceted and complex endeavor~\cite{arjovsky2017towards,radford2015unsupervised}.

\noindent
\textit{Self-training} presents an alternative approach to UDA, which has recently been shown to be highly efficient~\cite{zoph2020rethinking}.
It follows the principle that predictions generated in previous steps are used as pseudo labels for the next stage of network learning. 
The literature has addressed self-training for semantic segmentation in non-medical and medical applications and showed state-of-the-art performance on benchmark datasets \cite{zou2018unsupervised,zou2019,shin2020,nie2018asdnet,yu2019uncertainty,xia20203d}. 
A further categorization of methods for handling limited dataset annotations is available in the review by Tajbakhsh et al.~\cite{tajbakhsh2020embracing}.
The potential of integrating uncertainty guidance in self-training was recently demonstrated for segmenting sparsely annotated micro-CT scans~\cite{zheng2020cartilage}. 

\section{Methods}
\begin{figure}[t]
 \centering
 \includegraphics[keepaspectratio,width=\textwidth,height=\textheight]{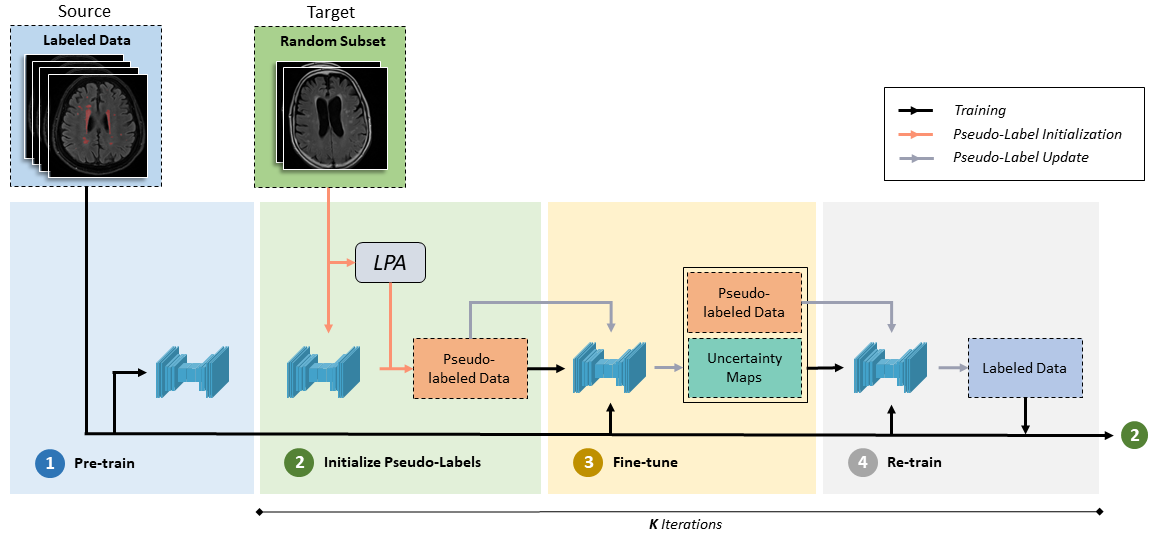}
 \caption{Illustration of our Self-Training pipeline for domain adaptation.}
 \label{fig:pipeline}
\end{figure}

\subsection{Problem Definition}
Given a labeled dataset from the source domain $S$ with samples $X^S = \{ {x_i^S}\}_{i = 1}^N$ and labels $Y^S = \{ {y_i^S}\}_{i = 1}^N$, and an unlabeled dataset from the target domain $T$ with samples $X^T = \{ {x_i^T}\}_{i = 1}^M$, the goal is to predict labels in the target domain. 
We want to achieve this goal by incorporating the large number of unlabeled target samples in the network training, where typically  $M > N$. 
In self-training, this is achieved by inferring pseudo target labels $\tilde{y}^T$ and updating them iteratively to improve the label quality and consequently the learning process.

\comment{Old: Prior to describing the proposed approach, we first detail the definitions of the self-training problem for UDA. The assumption holds that the framework operates with input samples $x \in X$, where $X$ is some input space and labels $y \in Y$ from label space $Y$. During training, the framework has full access to samples $x^S$ and the corresponding ground truth labels $y^S$ of source domain $S$. On the other hand, no ground truth labels $y^T$ exist for the target domain $T$. The framework has access, however, to a large number of samples $x^T$ of target domain $T$ with $\vert T \vert > \vert S \vert$. Source and target domains are similar and span different but overlapping regions of the input space $X$. Goal of the proposed approach is to use labeled samples of domain $S$ and unlabeled samples of domain $T$ to learn a predictive objective function $f: X \rightarrow Y$ which in turn accurately predicts labels $y^T$ for any input sample $x^T$. In self-training this is achieved by inferring pseudo target labels $y_p^T$ and updating them iteratively to improve the label quality and consequently the learning process.}

\subsection{STRUDEL: Self-Training with Uncertainty}
Fig.~\ref{fig:pipeline} provides a graphical overview of our proposed Self-TRaining with Uncertainty DEpendent Label refinement, with the pseudo-code in Algorithm~\ref{alg:pipeline}. 
First, we pre-train a base model on the source dataset $\left ( X^S, Y^S  \right )$ with standard supervised learning. 
The base model is then applied to a random subset (drawn without replacement) of the target sample, $r(X^T)$, of size $P$ to infer pseudo labels. 
However, these pseudo labels will initially not be of high quality due to the domain gap. 
Consequently, we propose to leverage existing WMH segmentation software, where we use LPA, to increase the quality of pseudo labels. 
To this end, we apply a pixel-wise OR operator between base model predictions and LPA prediction to obtain the  pseudo target labels $\tilde{Y}^T = \{ {\tilde y_i^T} \}_{i = 1}^{P}$. 

\noindent
In a third step, the base model is fine-tuned with $\tilde{Y}^T$. 
With this model, we segment again the same random sample drawn earlier, $r(X^T)$, producing pseudo labels of higher quality. Here, we assume that a model can generate better predictions than its noisy training labels~\cite{guan2018said}.
Next to the labels, we also infer the segmentation uncertainty $U$ at this stage. 
Finally, a new model is trained from scratch, where we add the updated pseudo labels $\tilde{Y}^T$ with the corresponding uncertainty $U$ to the training set. 
Training a new model at this point has advantages over fine-tuning as also noted in~\cite{zoph2020rethinking}. 
The labels inferred from this model on the random subset are then added to the fixed training set, which initially only consists of the annotated source data. 
We then continue with the next iteration in step 2, where this model serves as a base model, another random subset is sampled and pseudo labels are inferred. 

\begin{algorithm}[t]
\SetAlgoLined
 \SetKwInOut{Input}{input}
 \SetKwInOut{Output}{output}
 \Input{Source data $X^S$, Source labels $Y^S$, Target data $X^T$}
 \Output{Output model $\mM_K$}
 $r() \gets$ random sampler\;
 $\mM_0 \gets$ train base model with $(X^S, Y^S)$\;
  $D_{\text{fix}} \gets (X^S, Y^S)$ \tcp*[r]{initialize fixed training set}
 \For{$k\gets1$ \KwTo $K$}{
    $X^T_k \gets r(X^T)$ \tcp*[r]{sample random subset}
    $\tilde{Y}^T_{k} \gets pixel-wise\_or(\mM_{k-1}(X^T_k), LPA(X^T_k))$ \tcp*[r]{init. pseudo labels}
    $D_k \gets D_{\text{fix}} \cup (X^T_{k}, \tilde{Y}^T_{k})$ \tcp*[r]{merge training data}
    $\mM_{k-1} \gets$ fine-tune $\mM_{k-1}$ with $D_k$\;
    $\tilde{Y}^T_{k}, U_{k} \gets \mM_{k-1}(X^T_{k})$ \tcp*[r]{update labels and get uncertainty}
    $D'_k \gets D_{\text{fix}} \cup (X^T_{k}, \tilde{Y}^T_{k})$ \tcp*[r]{merge training data}
    $\mM_k \gets$ re-train model with $D'_k$ and uncertainty $U_{k}$\; 
    $D_{\text{fix}} \gets D_{\text{fix}} \cup (X^T_{k}, \mM_k(X^T_{k}))$ \tcp*[r]{update fixed training set}
 }
 \Return{$\mM_K$}\;
 \caption{Self-Training with Uncertainty on Noisy Labels}
 \label{alg:pipeline}
\end{algorithm}

\subsection{Uncertainty-Guided Pseudo Labels}
Inferring pseudo target labels usually has the disadvantage of label noise. 
To increase the robustness of our method against label noise, we propose an uncertainty-guidance that strengthens regions of low uncertainty and penalizes regions of high uncertainty. 
To estimate the uncertainty, we follow a Bayesian machine learning approach with an estimation of Monte Carlo (MC) samples by dropout~\cite{gal2015}. 
Accordingly, we train the backbone segmentation network with dropout layers and perform $C$ stochastic forward passes at test time to obtain Monte Carlo samples. The expectation over the MC samples $\mathbb{E}(\hat{y})$ provides us a more robust label prediction, which we use to update the pseudo-label.
Further, computing the variance across $C$ MC samples gives us a pixel-wise measure of uncertainty of the predicted segmentation:
\begin{equation}
   U(\hat{y})=\left \{ \sigma_1,...,\sigma_{H\times W} \right \} =  \frac{1}{C}\sum_{i = 1}^{C}\left (\widehat{y}_i - \mathbb{E}(\hat{y})) \right )^{2}, 
\end{equation}
where $U(\hat{y})$ denotes the uncertainty map, $\sigma_i$ the pixel-wise variance, $H,W$ the images height and width, and $\hat{y}_i$ the model prediction from the $i$th MC sample. 
Anticipating small values for the uncertainties, we rescale the values into the range [0, 1].
We integrate the uncertainty into network training by the definition of an uncertainty-aware binary cross entropy (UBCE) loss:
\begin{equation}
\mathcal{L}_{\text{UBCE}}=-\frac{1}{H\times W}\sum_{n=1}^{H\times W} (1-\sigma_n)\left [\tilde y_n\cdot log\left ( \hat{y}_n \right )+\left ( 1-\tilde y_n \right )\cdot log\left ( 1-\hat{y}_n \right )\right ].
\end{equation}
Note that the uncertainty-aware cross entropy $\mathcal{L}_{\text{UBCE}}$ is only applied to pseudo-labeled data within the re-training step (see~Algorithm~\ref{alg:pipeline} line 11), whereas the standard cross entropy $\mathcal{L}_{\text{BCE}}$ is applied to fixed data samples. The combined loss function is defined as:
\begin{equation}\label{loss}
    \mathcal{L}=\mathcal{L}_{\text{Dice}}+\mathcal{L}_{\text{BCE}} + \mathcal{L}_{\text{UBCE}}.
\end{equation}

\subsection{Segmentation Backbone Architectures}\label{architectures}
For the segmentation model $M$, we evaluate two neural network architectures. 
First, the U-Net \cite{ronneberger2015u}, which has proven its performance in difficult segmentation tasks and has been adapted and improved for various applications since then. 
Second, to explore the effects of a superior model architecture within the framework, we include a novel network architecture, the OctSE-Net, which introduces two modifications to the U-Net.
The first modification is to replace all convolution layers with octave convolutions~\cite{chen2019oct}, which factorize feature maps by their frequencies. Octave convolutions can increase segmentation performance, by offering a wider context, while reducing memory consumption. 
The second modification is to integrate squeeze \& excitation (SE) blocks~\cite{Hu_2018_CVPR}, more precisely the channel and spatial SE block~\cite{roy2019recalibrating}, which can boost accuracy by re-calibrating feature maps. 
Both modifications increase the receptive field without substantially increasing model parameters. 
This can improve the segmentation accuracy without encouraging overfitting and therefore help the generalization across domains. 
For uncertainty estimation, we insert dropout layers after each convolutional block in both architectures.


\section{Experiments and Results}
\subsection{Datasets}
As source dataset, we use data from the WMH challenge (\url{https://wmh.isi.uu.nl}), and, as target dataset, we use data from the Alzheimer's Disease Neuroimaging Initiative (ADNI) (\url{http://adni.loni.usc.edu}).
The WMH segmentation challenge dataset provides manual annotations for 60 subjects from 3 sites. 
For each subject, co-registered 3D T1-weighted and 2D multi-slice FLAIR scans are available, where we work with bias-field corrected T1 scans and original FLAIR scans as suggested by~\cite{hernandez2016computational,tubi2020white}.
The large ADNI-2 dataset~\cite{beckett2015} with over 3,000 scans from 58 sites serves as our multi-domain target data. 
For each subject, T1-weighted and 2D FLAIR scans are available, which we have linearly aligned within a session with ANTs~\cite{avants2009advanced}. T1 scans have further been bias field corrected with N4 normalization~\cite{tustison2010n4itk}, using the ANTs implementation. 
To quantitatively evaluate our methods on the ADNI dataset, we extracted a subset of 30 subjects based on scanner type and WMH lesion load for manual annotation\footnote{\label{note1}Code and manual segmentations will be made publicly available upon acceptance.}. 21 of these annotated scans serve solely as a test set and 9 are used to train alternative segmentation approaches, described in the next section.

\subsection{Implementation Details}
We implemented the proposed framework and all baseline experiments in Pytorch v.1.6.0\footnoteref{note1}.
Experiments were performed employing the Adam optimizer with default parameters (betas=(0.9, 0.999), eps=1e-08), learning rate 1e-4 and batch size 4.
Image intensities were normalized to zero-mean and unit-variance and axial slices center cropped to a consistent size of $192 \times 192$ pixels across all datasets.
Standard spatial augmentation techniques (flipping, rotation, scaling, and elastic transformation) were used during training for regularization.
In self-training, the size of the random subset per iteration is set to $P=35$. 
The thresholds to obtain binary segmentation maps for the creation of pseudo-labels were set to 0.5 and 0.75 for the network prediction and LPA, respectively. 
The high threshold for LPA mitigates hypersensitive responses.
We use 80 epochs for training from scratch and 20 epochs for fine-tuning.
We set the number of stochastic forward passes to $C = 10$. We found that increasing $C$ further does not improve the segmentation performance. 
The drop-out rate was set to 0.2.
No explicit post-processing was performed in any experiment.
A Geforce Titan RTX GPU was predominantly used for training and testing. 

\subsection{Experiments}
We perform several experiments to evaluate the domain transfer performance of different approaches.
First, as \textbf{Base Model}, we directly apply a model trained on the source data on the target domain data without any adaptation. 
Next, we evaluate two approaches that use a small labeled subset of the target domain during training. The \textbf{Joint Model} combines the target and source training data, and the \textbf{Fine-Tuning} model uses the labeled target data to fine-tune the Base Model. 
Finally, we evaluate two UDA approaches with pseudo labels: \textbf{Self-Training} without uncertainty guidance and the proposed \textbf{STRUDEL} with uncertainty guidance, both using LPA labels. 
We also report results for STRUDEL without LPA and for LPA itself, where we set the threshold parameter to $0.45$, as suggested  in~\cite{vanderbecq2020} for  ADNI.
We evaluate the segmentation accuracy for all experiments by following evaluation metrics suggested by the WMH segmentation challenge~\cite{kuijf2019standardized}: (1)  Dice Similarity Coefficient (DSC), (2)  modified Hausdorff distance (95th percentile; H95), (3) absolute log-transformed volume difference (lAVD), (4) sensitivity for detecting individual lesions (Recall), and (5) F1-score for individual lesions (F1).

\subsection{Results \& Discussion}
\begin{figure}[t]
 \centering
 \includegraphics[keepaspectratio,width=\textwidth,height=\textheight]{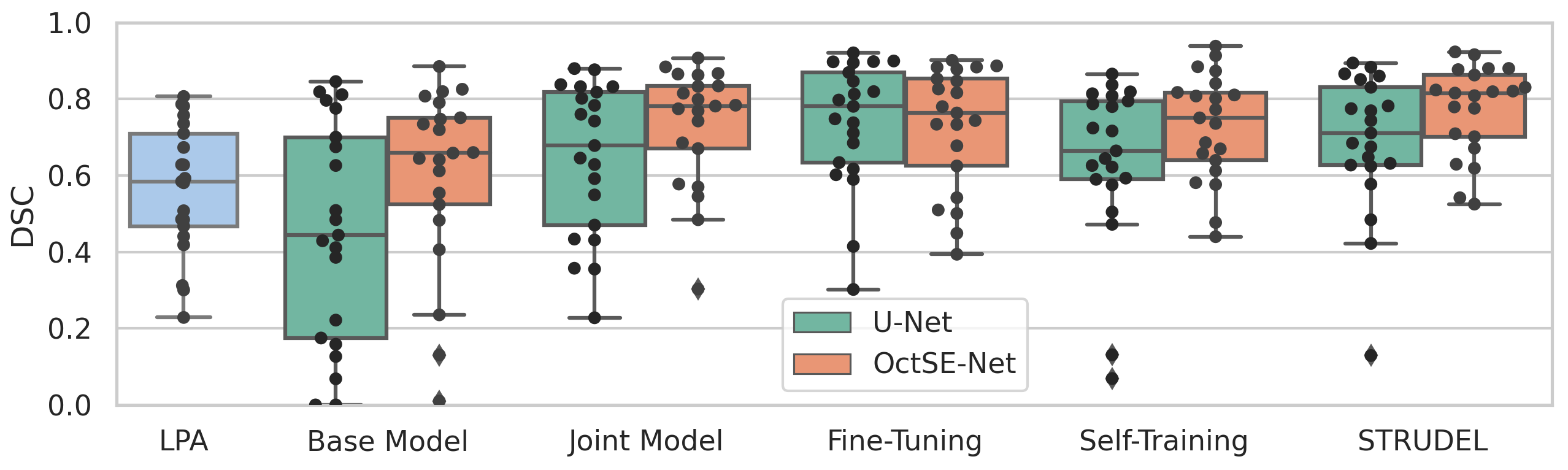}
 \caption{Boxplot of Dice Similarity Coefficient for the different methods. 
 Points outside the whiskers are determined as outliers based on the inter-quartile range.}
 \label{fig:boxplot}
\end{figure}

\begin{table}[t]
    \centering
    \caption{Comparison of segmentation methods, network architectures, type of used data (S: Source manual labels, T: Target manual labels, P: target pseudo labels), and their mean performance $\pm$ standard deviation on the metrics: Dice Coefficient (DSC), 95th Percentile Hausdorff Distance (H95), log transformed absolute volume difference (lAVD), lesion Recall and F1.}
    \ra{1.2}
    \resizebox{\linewidth}{!} {
    \begin{tabular}{@{} l c c c c@{\hskip 5\tabcolsep} c@{\hskip 5\tabcolsep} c @{\hskip 5\tabcolsep}c @{\hskip 5\tabcolsep}c@{\hskip 5\tabcolsep} c@{}}
    \toprule
     Methods & S & T & P& DSC $\uparrow$ & H95 [mm] $\downarrow$ & lAVD $\downarrow$ & Recall $\uparrow$ & F1 $\uparrow$\\
    \midrule
         LPA & \xmark&\xmark&\xmark& 0.57$\pm$0.16 &   23.1$\pm$23.4 & 0.71$\pm$0.49 &    0.81$\pm$0.16    &  0.39$\pm$0.18   \\
         \hdashline
         \multicolumn{10}{c}{U-Net}\\
         \midrule
         Base Model   &\cmark&\xmark&\xmark& 0.45$\pm$0.28 &  27.1$\pm$37.5  & 1.09$\pm$1.70 & 0.67$\pm$0.32 & 0.48$\pm$0.21  \\ 
         Joint Model  & \cmark&\cmark&\xmark& 0.64$\pm$0.19 & 17.2$\pm$25.0  & 0.60$\pm$0.52 & 0.74$\pm$0.29 & 0.52$\pm$0.15  \\ 
         Fine-Tuning & \cmark&\cmark&\xmark& \ubold{0.73$\pm$0.16} & \ubold{11.2$\pm$23.0}  & 0.36$\pm$0.41 & \ubold{0.75$\pm$0.22} & \ubold{0.65$\pm$0.14} \\
         Self-Training  & \cmark&\xmark&\cmark& 0.64$\pm$0.20 & 17.8$\pm$28.8 & 0.51$\pm$0.68 & 0.51$\pm$0.27 & 0.50$\pm$0.23 \\
         STRUDEL  & \cmark&\xmark&\cmark& 0.69$\pm$0.18 & \ubold{11.2$\pm$14.5}  & \ubold{0.30$\pm$0.32} & 0.58$\pm$0.27 & 0.64$\pm$0.22 \\
         \multicolumn{10}{c}{OctSE-Net}\\
         \midrule

         Base Model &\cmark&\xmark&\xmark& 0.60$\pm$0.23 & 19.7$\pm$29.5 & 0.77$\pm$1.12  &  0.80$\pm$0.26 &  0.61$\pm$0.19  \\ 
         Joint Model  & \cmark&\cmark&\xmark& 0.73$\pm$0.15 & 11.8$\pm$24.7  & 0.34$\pm$0.37 & \ubold{0.89$\pm$0.10} & 0.59$\pm$0.14 \\
         Fine-Tuning& \cmark&\cmark&\xmark&  0.73$\pm$0.15 & 11.4$\pm$23.4 & 0.41$\pm$0.38 & 0.77$\pm$0.18 & 0.64$\pm$0.17 \\
         Self-Training & \cmark&\xmark&\cmark&  0.73$\pm$0.13 & 14.7$\pm$18.2 & \ubold{0.25$\pm$0.27} & 0.56$\pm$0.21 & 0.63$\pm$0.17 \\
         STRUDEL & \cmark&\xmark&\cmark&   \ubold{0.78$\pm$0.10} & \ubold{7.79$\pm$8.52} & 0.27$\pm$0.23 & 0.77$\pm$0.16 & \ubold{0.70$\pm$0.15} \\
         $\hookrightarrow$ w/o LPA & \cmark&\xmark&\cmark&   0.67$\pm$0.20 & 12.9$\pm$13.4 & 0.63$\pm$0.58 & 0.58$\pm$0.23 & 0.66$\pm$0.18 \\
\bottomrule
    
    \end{tabular}
    }
    \label{results}
\end{table}{}
Table~\ref{results} reports the quantitative segmentation results on the ADNI target domain. 
Fig.~\ref{fig:boxplot} shows the DSC in more details as boxplots. 
As a reference, the DSC of the Base Models on the source dataset are 0.73 for U-Net and 0.76 for OctSE-Net.
We observe that the direct transfer of the Base Model on the ADNI dataset performs poorly, regardless of the backbone architecture. 
LPA beats the baseline U-Net by a large margin, which is in accordance with the results described in~\cite{vanderbecq2020}.
OctSE-Net outperforms LPA, which confirms our assumption that OctSE-Net is a more robust architecture. 
However, the more detailed results in Fig.~\ref{fig:boxplot} show that both base models produce some predictions with zero DSC, whereas LPA does not. 
These outliers can lead to poorly initialized pseudo labels, which is the reason for including LPA in pseudo-label initialization, confirmed by the bad results for STRUDEL w/o LPA.
We report results for all self-training-based experiments after 5 iterations, as no further improvement was observed afterwards.
STRUDEL outperforms all other methods in DSC and H95, and is best or second best in lAVD and lesion F1. 
LPA and the joint model perform best in terms of lesion recall.
Both of these methods have relatively poor performance in lesion F1, which is the result of a high number of false positive predictions. 
STRUDEL performs strongly in both metrics, which we believe is due to the uncertainty capturing false positives reliably.
Results of a Wilcoxon signed-rank test on DSC show that the improvement of STRUDEL with respect to Self-Training is significant for OctSE-Net ($p<0.005$) and also that the improvement of OctSE-Net with respect to U-Net is significant for STRUDEL ($p<0.001$).

%

\section{Conclusion}
Self-Training is a simple and effective approach for UDA, however noisy pseudo labels can limit its effectiveness. 
In this work, we proposed STRUDEL, an uncertainty-guided self-training method for unsupervised domain adaptation.
We found that introducing uncertainty into the objective function can efficiently guide the learning process in the presence of noisy labels.
We further demonstrated that leveraging an existing algorithm (LPA) for pseudo label initialization can additionally boost performance.
Our experimental results showed that Self-Training with uncertainty guidance is a strong approach for UDA, which in combination with a strong and robust network architecture, can even outperform supervised methods.



\section*{Acknowledgments}
This research was partially supported by the Bavarian State Ministry of Science and the Arts and coordinated by the bidt, and the BMBF  (DeepMentia,031L0200A).

\comment{\begin{figure}[h]
 \centering
 \includegraphics[keepaspectratio,width=\textwidth,height=\textheight]{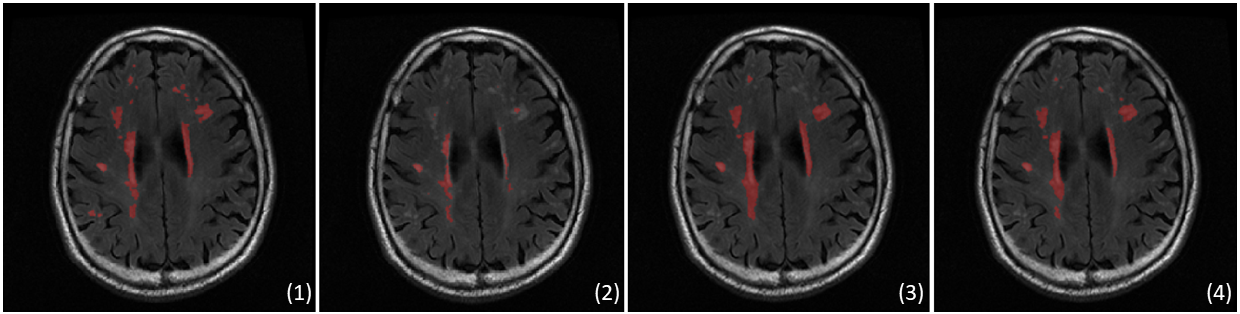}
 \caption{Segmentation maps overlaid on FLAIR scan for (1) ground truth, (2) Inital pseudo label, (3) Self-Training, (4) STRUDEL}
 \label{fig:results}
\end{figure}
}
%
%
%
\bibliographystyle{splncs04}
\bibliography{references}
%
\end{document}